\documentclass[10pt,twocolumn,letterpaper]{article}

\usepackage{btas}
\usepackage{times}
\usepackage{epsfig}
\usepackage{graphicx}
\usepackage{amsmath}
\usepackage{amssymb}
\DeclareMathOperator*{\argmin}{arg\,min}
\graphicspath{{./images/}}
\usepackage{url}
\usepackage{subfig}
\usepackage{fancyhdr}
 \usepackage{float}
\usepackage{multirow}

\btasfinalcopy 

\ifbtasfinal\fi

\begin{document}

\title{Learning A Shared Transform Model for Skull to Digital Face Image Matching}


\author{Maneet Singh$^1$, Shruti Nagpal$^1$, Richa Singh$^{1}$, Mayank Vatsa$^{1}$, and Afzel Noore$^2$\\
$^1$IIIT-Delhi, India, $ ^2$Texas A\&M University, Kingsville, USA\\
\{\tt\small maneets, shrutin, rsingh, mayank\}@iiitd.ac.in, afzel.noore@tamuk.edu
}

\maketitle
\thispagestyle{fancy}
\fancyhf{}
\renewcommand{\headrulewidth}{0pt}

\begin{abstract}
Human skull identification is an arduous task, traditionally requiring the expertise of forensic artists and anthropologists. This paper is an effort to automate the process of matching skull images to digital face images, thereby establishing an identity of the skeletal remains. In order to achieve this, a novel Shared Transform Model is proposed for learning discriminative representations. The model learns robust features while reducing the intra-class variations between skulls and digital face images. Such a model can assist law enforcement agencies by speeding up the process of skull identification, and reducing the manual load. Experimental evaluation performed on two pre-defined protocols of the publicly available IdentifyMe dataset demonstrates the efficacy of the proposed model.

\end{abstract}

\section{Introduction}
\begin{figure}
\centering
\includegraphics[width = 3.3in] {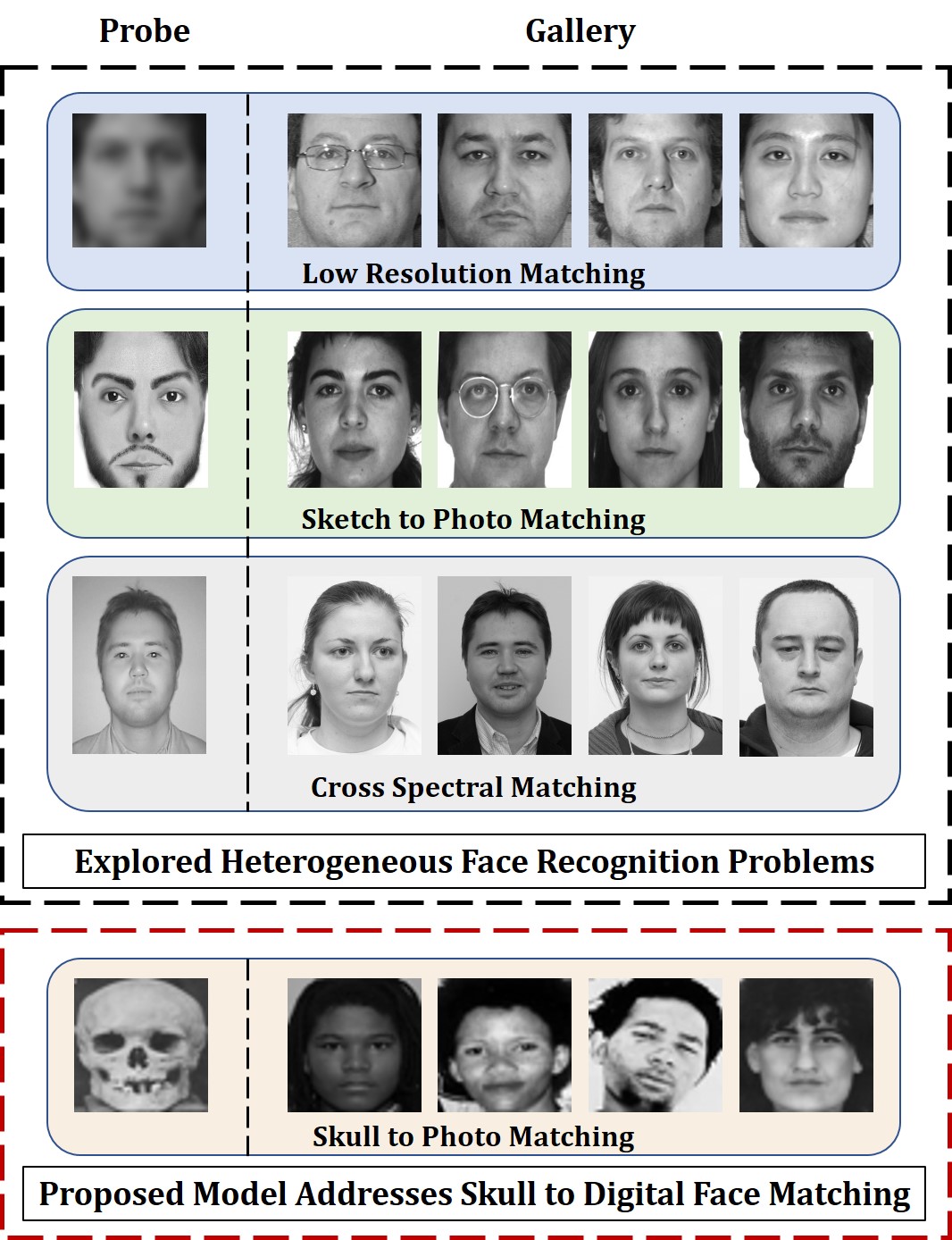}
\caption{In literature, heterogeneous matching problems such as low resolution, sketch face recognition, and visible to near infrared (cross spectral) face recognition have received substantial attention. This research addresses the less explored heterogeneous problem of skull to digital face image matching.}
\label{fig:intro}
\end{figure}

Human skull identification is a challenging problem of vital importance in forensics and law enforcement, which involves matching a skull image with gallery face images (Figure \ref{fig:intro}). Unidentified skulls often result in unrest with the law enforcement officials and family of the victim. For instance, in 2014, skeletal remains (including a skull) were found in North Creek, just above Virginia \cite{news2014}. Forensic experts were able to predict an age-range, gender, and ethnicity of the skull, however, the identity remains unknown. Efforts were made to investigate missing person reports around that area, and other unsolved cases. Retired police chief, Kurt Wright, who worked on the case, mentioned ``This has always just kind of nagged at me for some reason, and I wanted some closure to it'' \cite{news2014}. In 2017, the case was again publicized in news to look for fresh leads; however, the case remains unsolved. 



\begin{figure*}
\centering
\includegraphics[width = 5in] {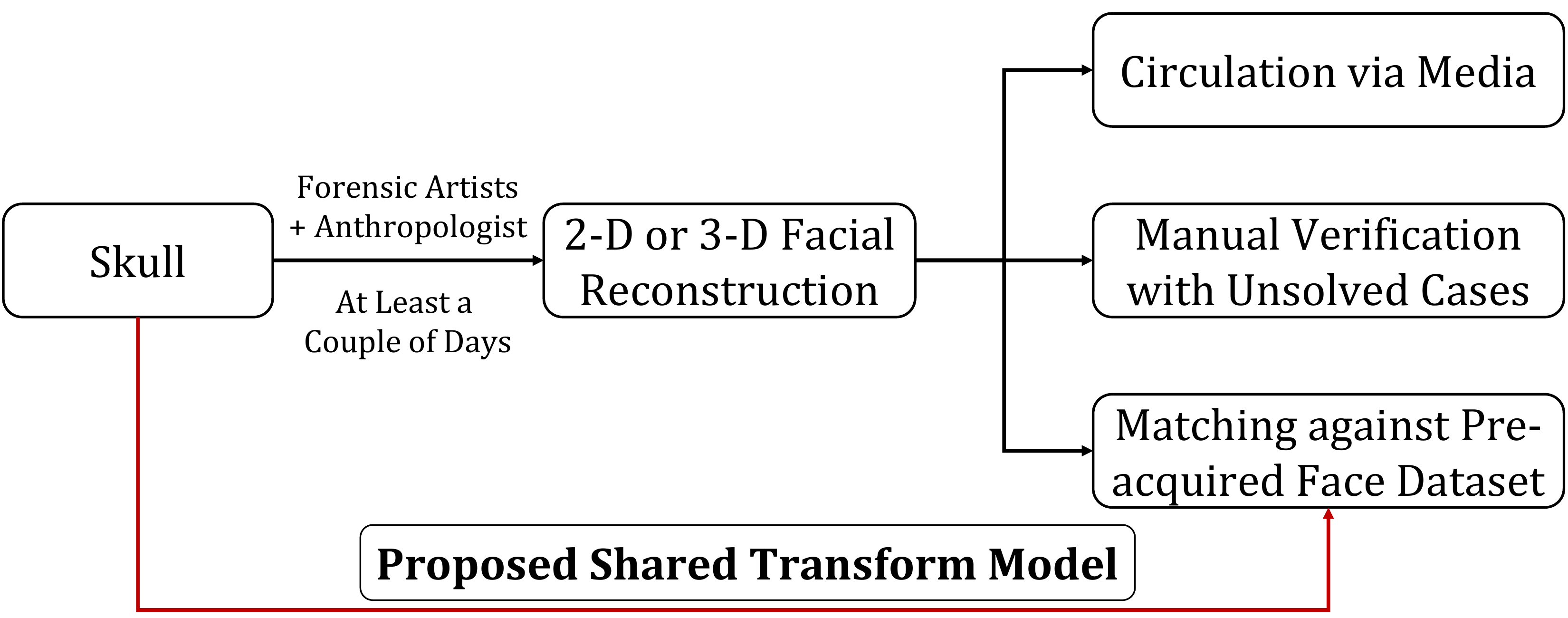}
\vspace{5pt}
\caption{Traditional techniques utilize facial reconstruction, a time consuming process, often requiring the input of inter-disciplinary experts. The proposed Shared Transform Model eliminates the need for reconstruction by directly matching a skull image with a dataset of face images.}
\label{motivation}
\end{figure*} 

As shown in Figure \ref{motivation}, skull identification is primarily performed by means of a manual reconstruction approach. That is, given a skull image, a facial reconstruction is created which is then circulated via digital or print media, or matched with a given gallery of digital face images. The techniques for facial reconstruction can broadly be divided into three categories: (i) 2-D reconstruction, (ii) 3-D reconstruction, and (iii) superimposition. 2-D reconstruction techniques construct a 2-D face image from a given skull image, whereas 3-D reconstruction constructs a 3-D face, using either Computer Graphics or via clay or sculptures. Superimposition refers to the task of superimposing a digital face image over a skull, and analyzing its alignment accuracy. This technique is generally applied when the forensic experts have a small subset of possible identities to be verified against a given skull. Traditionally, the successful application of the above techniques require at least a couple of days to a couple of weeks, along with collaboration between forensic artists and anthropologists. In this research, we propose to automate the process of skull identification. If a completely intact skull is found, it may enable the law enforcement agencies to determine its identity without performing the facial reconstruction. With the availability of a skull recognition algorithm, the search for the identity of a skull can extend beyond the immediate location in a fast, effective, and automated manner. 

Skull identification can be viewed as a heterogeneous matching problem, where one domain corresponds to skull images, while the second contains digital face images. Figure \ref{fig:intro} shows different heterogeneous face matching problems explored in literature along with a sample skull to digital face matching scenario. It can be observed that the information content of digital face images varies significantly from that of the skull image. Heterogeneous face matching has received significant attention in literature, in terms of low resolution face matching, or sketch to face matching, or visible to near infrared (NIR) face matching \cite{nagpal2016sketch, surveyHFR}. However, the task of automated skull recognition has received limited attention \cite{survey95, survey2011}. Most of the existing algorithms focus on the task of facial reconstruction \cite{survey15} by means of superimposition, that is, scenarios having the availability of a small subset of possible identities for a given skull. Besides reconstruction, researchers have also focused on automating the task of matching 3-D CT head scans with a gallery of digital images \cite{ct2, ct1}. While this is a viable option to perform matching, it incurs the overhead charge of conducting a CT head scan, and the inherent assumption that the skull found can be used for scanning. A model for automated skull recognition without the need of reconstructions or scans has been proposed by Nagpal \textit{et al.} \cite{ijcbSkull}. The authors also proposed a publicly available IdentifyMe dataset consisting of 464 skull images, along with Semi-supervised transform learning (SS-TL) and Unsupervised transform learning (US-TL) models for skull to digital face image matching. 

\begin{figure*} 
\centering
\includegraphics[width = 7in] {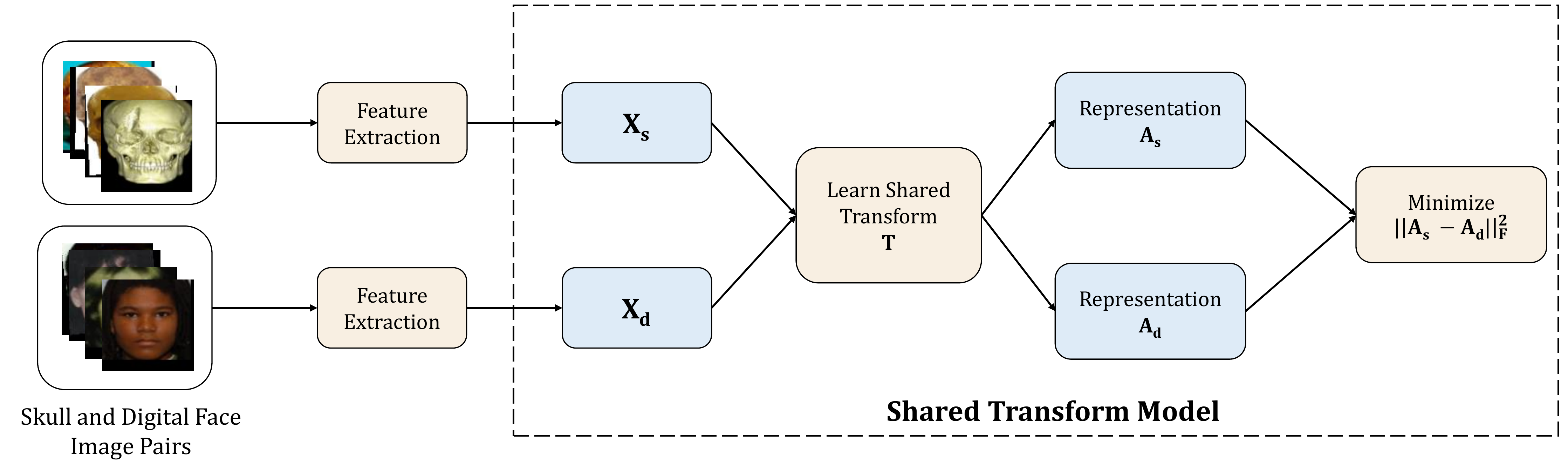}
\caption{Diagrammatic representation of the proposed Shared Transform Model. Given mated pairs of skull and digital face image ($\mathbf{X_s}$ and $\mathbf{X_d}$), the proposed algorithm learns a shared transform matrix $\mathbf{T}$, along with the corresponding representations $\mathbf{A_s}$ and $\mathbf{A_d}$. The model is optimized in order to reduce the intra-class variations of the given training data.}
\label{algo}
\end{figure*}

Existing algorithms for skull recognition rely either on the availability of CT head scans or require learning a large number of parameters. Given the availability of limited training samples, this might result in over-fitting and thus lower the identification performance on unseen probes. In order to address these limitations, this research proposes a novel Shared Transform Model (Figure \ref{algo}), which learns a single ``shared transform'' for both skulls and digital face images, while reducing the intra-class variations. Experimental results on the publicly available skull dataset, IdentifyMe \cite{ijcbSkull} demonstrate the efficacy of the proposed model for the task of skull to face matching.


\begin{figure}
\centering
\includegraphics[width = 3.3in] {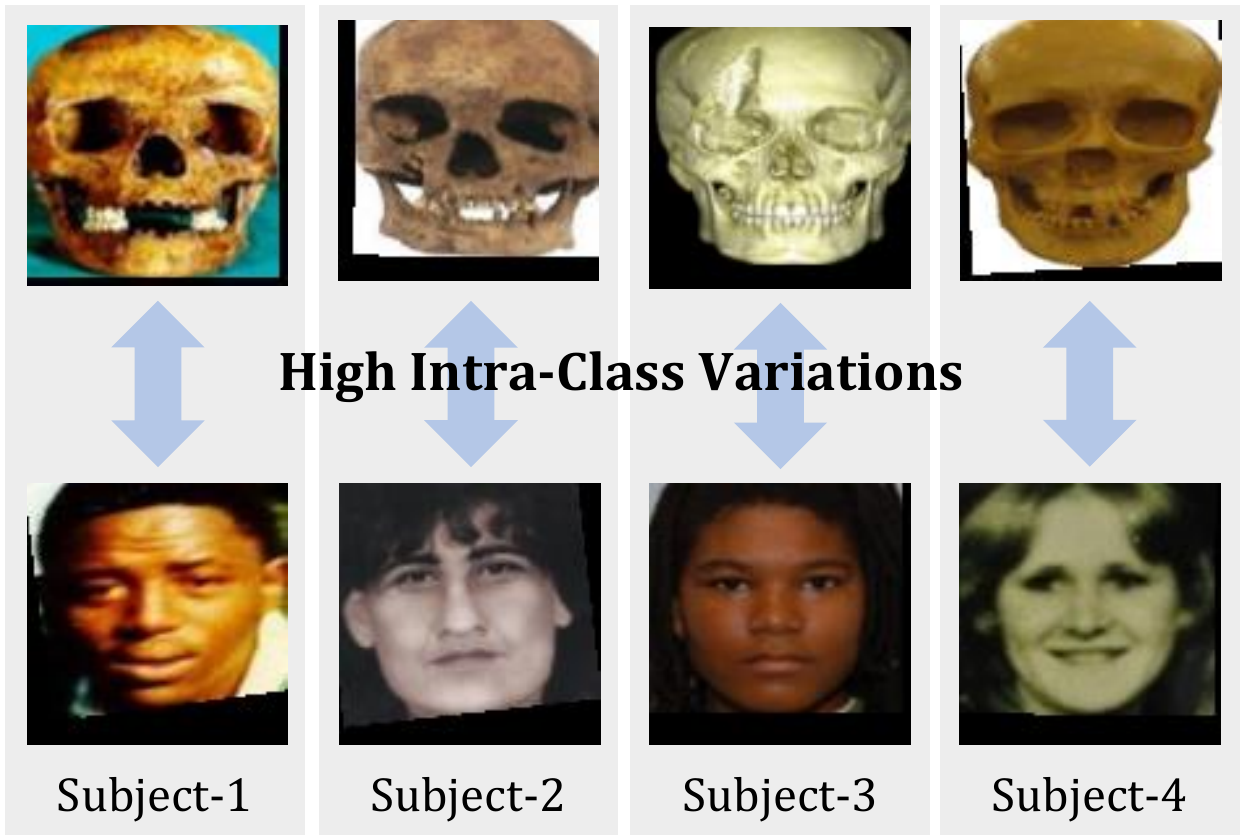}
\caption{Sample images of four subjects of the IdentifyMe dataset. The top row corresponds to the skull images, while the bottom rows contains the digital face images. Large variations can be observed across the two images of a particular subject. }
\label{intraClass}
\end{figure}

\section{Proposed Shared Transform Model}
Figure \ref{intraClass} presents sample skull and digital face image pairs. It can be observed that the images corresponding to a given subject showcase large variations in terms of the information content. Coupled with limited training samples, high intra-class variations render the problem of skull to digital face image matching challenging. Existing research in skull to photo matching has utilized transform learning \cite{ijcbSkull}. The superior performance of transform learning models for other heterogeneous tasks with limited training samples \cite{nagpalIccv} further make it an appropriate choice for addressing the given problem. However, the models proposed by Nagpal \textit{et al.} \cite{ijcbSkull} for the given task, SS-TL and US-TL, require learning a lot of parameters and include separate transforms for skulls and faces. In order to reduce the number of parameters and improve the overall recognition performance, in this research, a Shared Transform Model is proposed for learning representations such that the intra-class variations are minimized. Learning a shared transform reduces the number of parameters to be learned, and minimizing the intra-class variations incorporates discriminability in the proposed model.  

Transform Learning algorithms \cite{transform2, transform} learn a transform matrix $\mathbf{T}$ for an input $\mathbf{X}$, in order to generate a sparse representation $\mathbf{A}$. Mathematically, this is represented as:
\begin{equation} \label{TL}
\begin{gathered}
\min_{\mathbf{T}, \mathbf{A}} \ \|\mathbf{TX} - \mathbf{A}\|_{F}^{2} \ + \ \lambda_1\|\mathbf{T}\|_{F}^{2} \ + \ \lambda_2\log det(\mathbf{T}) \\
s.t. \ \|\mathbf{A}\|_0 \leq \tau
\end{gathered}
\end{equation}
where, the second and the third terms correspond to regularizers for ensuring a non-degenerate and scale balanced transform matrix. $\lambda_1$ and $\lambda_2$ control the weight given to the regularizers while optimizing the entire loss function. The sparsity constraint enforces the model to learn representative latent variables, thereby extracting meaningful information from the given data. 

The proposed Shared Transform Model learns representations such that the features of skull and digital face images are projected onto a common space, while reducing the intra-class variations. A shared transform is learned by projecting the given data onto a common space. That is, using the training data: skulls ($\mathbf{X_s}$) and digital face images ($\mathbf{X_d}$), a shared transform ($\mathbf{T}$) is learned as follows: 

\begin{equation} \label{sharedTL}
\begin{gathered}
\min_{\mathbf{T}, \mathbf{A_d}, \mathbf{A_s}} \ \|\mathbf{TX_s} - \mathbf{A_s}\|_{F}^{2} \ + \ \|\mathbf{TX_d} - \mathbf{A_d}\|_{F}^{2} \ \\ 
+ \ \lambda_1\|\mathbf{T}\|_{F}^{2} + \lambda_2\log det(\mathbf{T}) \ \\
s.t. \  (\|\mathbf{A_s}\|_0 \leq \tau \ , \ \|\mathbf{A_d}\|_0 \leq \tau)
\end{gathered}
\end{equation}

Equation \ref{sharedTL} learns a shared transform matrix $\mathbf{T}$ and sparse representations ($\mathbf{A_s}$, $\mathbf{A_d}$) for skull and digital face images, respectively. This projects the input data onto a common feature space; however, it does not ensure that the intra-class variations are minimized. In order to obtain features that reduce the within-class variations, the proposed Shared Transform Model incorporates a distance minimizing term between the representations as follows:

\begin{equation} \label{sharedTLFinal}
\begin{gathered}
\min_{\mathbf{T}, \mathbf{A_d}, \mathbf{A_s}} \ \|\mathbf{TX_s} - \mathbf{A_s}\|_{F}^{2} \ + \ \|\mathbf{TX_d} - \mathbf{A_d}\|_{F}^{2} \ \\ 
+ \ \lambda_1\|\mathbf{T}\|_{F}^{2} + \lambda_2\log det(\mathbf{T}) \ + \ \lambda_3\|\mathbf{A_d} - \mathbf{A_s}\|_{F}^{2}\\
s.t. \  (\|\mathbf{A_s}\|_0 \leq \tau \ , \ \|\mathbf{A_d}\|_0 \leq \tau)
\end{gathered}
\end{equation}

Similar to Equation \ref{sharedTL}, the above objective function learns a shared transform matrix $\mathbf{T}$, along with the representations of skull and digital face images ($\mathbf{A_s}$, $\mathbf{A_d}$). The additional distance minimizing term, ($\|\mathbf{A_d} - \mathbf{A_s}\|_{F}^2$) promotes learning representations such that the distance between them is minimized. Learning a shared transform matrix projects the heterogeneous input data onto a common feature space, while the distance minimizing term facilitates learning of discriminative features. Given mated pairs of skull and digital face images, the proposed Shared Transform Model can be learned via a standard alternating minimization approach which alternates over the variables to be learned for a given model. For the proposed model, this is achieved by iterating over a three step approach given below:

\begin{enumerate}
\item Optimize for $\mathbf{T}$:
\begin{equation}\label{optForT}
\begin{gathered}
\min_{\mathbf{T}} \ \|\mathbf{TX_s} - \mathbf{A_s}\|_{F}^{2} \ + \ \|\mathbf{TX_d} - \mathbf{A_d}\|_{F}^{2} \ \\ 
+ \ \lambda_1\|\mathbf{T}\|_{F}^{2} + \lambda_2\log det(\mathbf{T}) 
\\ \equiv\ \min_{\mathbf{T}} \left \| \mathbf{T} \binom{\mathbf{X_s}} {\mathbf{X_d}} - \binom{\mathbf{A_s }} {\mathbf{A_d}} \right \|_{F}^{2} + \ \lambda_1\|\mathbf{T}\|_{F}^{2} \\
+ \lambda_2\log det(\mathbf{T}) 
\end{gathered}
\end{equation}
The above equation is of the form given in Equation \ref{TL}, and is thus optimized with a closed form solution explained earlier by Ravishankar \textit{et al.} \cite{transform}.
\item Optimize for $\mathbf{A_s}$:
\begin{equation} \label{optForAs}
\begin{gathered}
\min_{\mathbf{A_s}} \ \|\mathbf{TX_s} - \mathbf{A_s}\|_{F}^{2} \ + \ \lambda_3\|\mathbf{A_d} - \mathbf{A_s}\|_{F}^{2}\\
s.t. \  \|\mathbf{A_s}\|_0 \leq \tau \\
\\ \equiv\ \min_{\mathbf{A_s}} \left \| \binom{\mathbf{TX_s}} {\mathbf{\sqrt{\lambda_3}A_d}} - \binom{\mathbf{I}} {\mathbf{I}}\mathbf{A_s} \right \|_{F}^{2} s.t. \  \|\mathbf{A_s}\|_0 \leq \tau \\  
\end{gathered}
\end{equation}

\item Optimize for $\mathbf{A_d}$:
\begin{equation} \label{optForAd}
\begin{gathered}
\min_{\mathbf{A_d}} \ \|\mathbf{TX_d} - \mathbf{A_d}\|_{F}^{2} \ + \ \lambda_3\|\mathbf{A_d} - \mathbf{A_s}\|_{F}^{2}\\
s.t. \  \|\mathbf{A_d}\|_0 \leq \tau \\
\equiv\ \min_{\mathbf{A_d}} \left \| \binom{\mathbf{TX_d}} {\mathbf{\sqrt{\lambda_3}A_s}} - \binom{\mathbf{I}} {\mathbf{I}}\mathbf{A_d} \right \|_{F}^{2} s.t. \  \|\mathbf{A_d}\|_0 \leq \tau \\  
\end{gathered}
\end{equation}
The above two equations are of the standard form given in Equation \ref{TL}, and thus have a closed form solution provided by Ravishankar \textit{et al.} \cite{transform}. 
\end{enumerate} 
Given mated pairs of skulls and digital face images, the above three steps are optimized iteratively to obtain the final transform matrix and corresponding representations.

\subsection{Identification using Shared Transform Model}
The training data is used to learn the shared transform matrix ($\mathbf{T}$), which is utilized at the test time to perform identification. In a real world scenario, the task is to match a given probe skull image ($x_{Skull}$) against a gallery of digital face images ($\mathbf{X_{Gal}}$). Here, $\mathbf{X_{Gal}}$ contains the gallery images arranged in a column wise manner, such that $x_{Gal}^{i}$ corresponds to the $i^{th}$ gallery image. Feature extraction is performed for all the gallery images using the transform matrix $\mathbf{T}$:
\begin{equation} \label{testGal}
\mathbf{A_{Gal}} = \mathbf{T}\times\mathbf{X_{Gal}} \ s.t. \ \|\mathbf{A_{Gal}}\|_0 \ \leq \tau
\end{equation}
Similarly, the representation of a given test image ($x_{Skull}$) is calculated as follows: 
\begin{equation} \label{testProbe}
a_{Skull} = \mathbf{T}\times{x_{Skull}} \ s.t. \ \|{a_{Skull}}\|_0 \ \leq \tau
\end{equation}
This is followed by calculating the Euclidean distance between the probe feature and the gallery features, in order to obtain the identity having the least distance:
\begin{equation} \label{testProbe}
\argmin_{i} \|a_{Skull} - a_{gal}^{i} \|_{2}^{2}
\end{equation}
The identity of $x_{gal}^{i}$ is returned as the identity of the given probe.

\begin{figure}
\centering
\includegraphics[width = 3in] {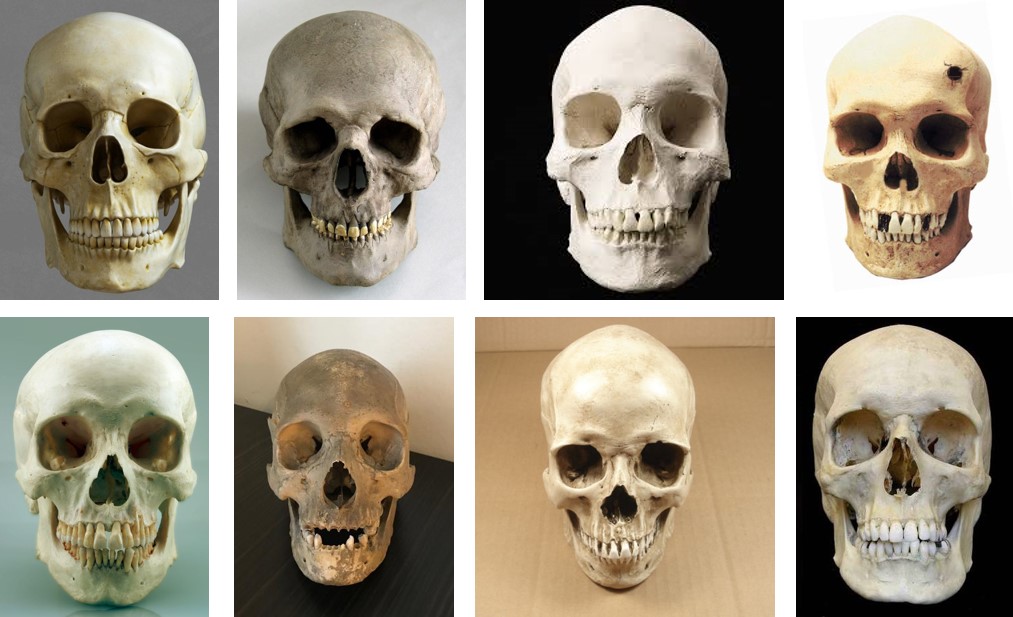}
\caption{Sample skull images of the IdentifyMe dataset.}
\label{fig:db}
\end{figure}

\begin{figure*} [t]
\centering
\subfloat[][Protocol-1]{\includegraphics[width = 3.5in] {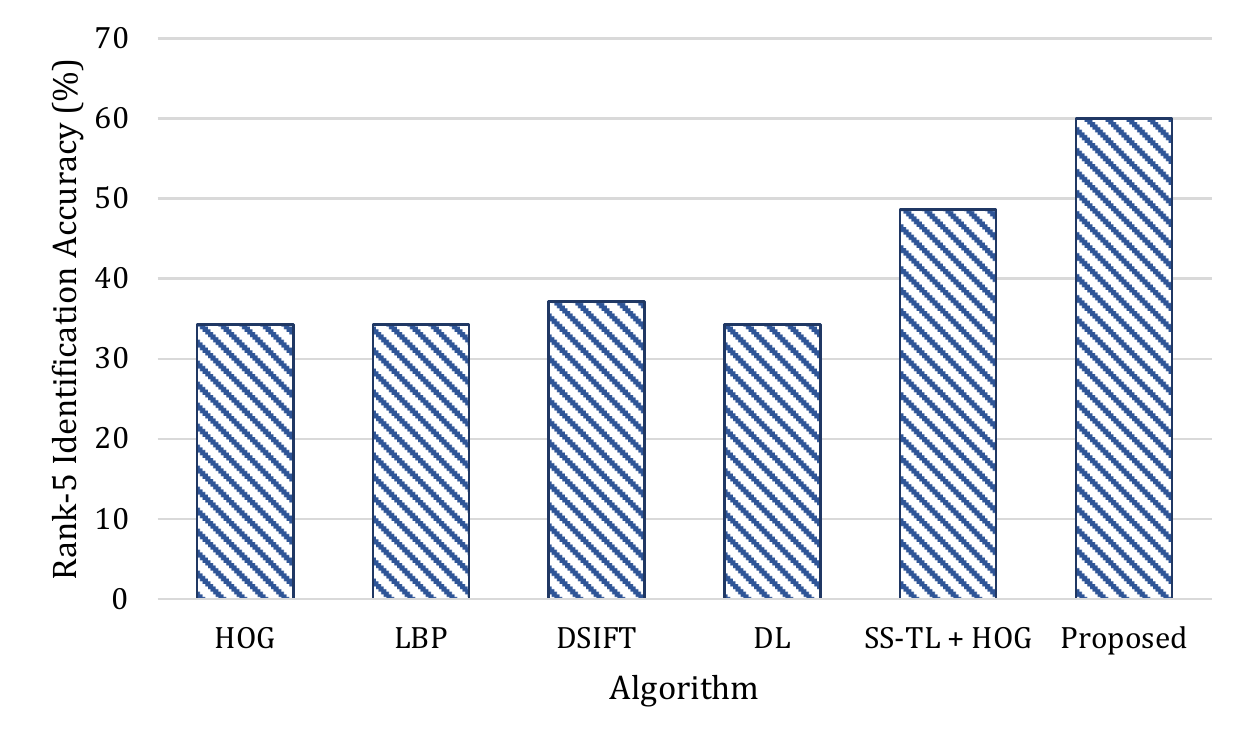}} 
\subfloat[][Protocol-2]{\includegraphics[width = 3.5in] {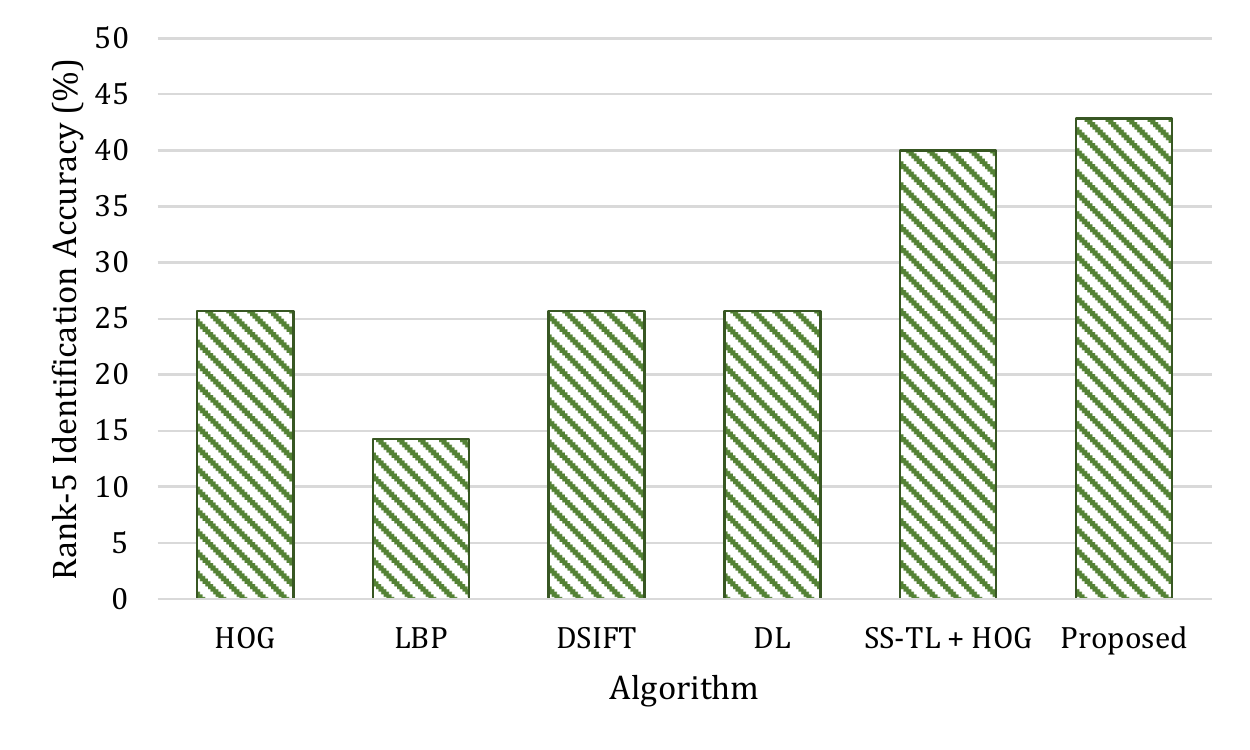}}
\caption{Bar graph representing the rank-5 identification accuracy for both the protocols on IdentifyMe dataset. SS-TL+HOG corresponds to the current state-of-the-art approach \cite{ijcbSkull}.}
\label{bar}
\vspace{-10pt}
\end{figure*}

\section{Experiments and Results}
There exists only one publicly available dataset for performing skull to digital face image matching, the IdentifyMe dataset \cite{ijcbSkull}. The dataset contains 464 skull images, inclusive of 35 mated skull images and their corresponding face images. Figure \ref{fig:db} presents sample images of the IdentifyMe dataset. The authors have also provided two protocols and baseline results for evaluating the performance of any algorithm. The two protocols are as follows:
\begin{itemize}
\item \textbf{Protocol-1}: This protocol evaluates the performance of a given algorithm on the 35 mated pairs of skull and digital face image. The results are reported with five fold cross validation. 
\item \textbf{Protocol-2 (extended gallery)}: This protocol attempts to mimic a real world scenario where the gallery consists of more subjects as compared to those present in the dataset. An extended gallery of 993 subjects is used for evaluation, such that for a given fold, the gallery contains 1000 subjects (993 extended and 7 from the IdentifyMe dataset) at test time. Similar to the previous protocol, five fold cross validation is performed for evaluation. 
\end{itemize}

The proposed Shared Transform Model is evaluated on both the protocols. Face images of IdentifyMe dataset are detected using Viola Jones face detector \cite{viola}, followed by geometric alignment with the skull images. Data augmentation is performed by combinations of flipping across the y-axis and modifying the brightness or contrast of images. Features are extracted from the pre-processed images, which are then used as input for the proposed Shared Transform Model. The learned model is then used with the test images, along with Euclidean distance for performing matching with the final transformed features (as shown in Section 3.1). 

Baseline results with Histogram of Oriented Gradients (HOG) \cite{hog}, Local Binary Pattern (LBP) \cite{lbp}, Dense Scale Invariant Feature Transform (DSIFT) \cite{dsift}, and Dictionary Learning (DL) \cite{dictionary} have been reported on the two protocols of IdentifyMe dataset. Since the best baseline performance is obtained with Histogram of Oriented Gradients \cite{hog} features, the proposed Shared Transform Model is applied on the same. Table \ref{results} and Figure \ref{bar} present the rank-1 and rank-5 identification accuracy (\%) for both the protocols. The key results are as follows:

\begin{table}
\centering
\caption{Rank-1 identification accuracy (\%) on the IdentifyMe dataset. Baseline results have directly been taken from \cite{ijcbSkull}.}
\label{results}
\begin{tabular}{|l|l|c|c|}
\hline
\multicolumn{2}{|l|}{\textbf{Algorithm}} & \textbf{Protocol-1} & \textbf{Protocol-2} \\
\hline
\hline
 \parbox[t]{2mm}{\multirow{6}{*}{\rotatebox[origin=c]{90}{Baseline}}} & LBP \cite{lbp} & 28.5 & 14.2\\
\cline{2-4}
& DSIFT \cite{dsift} & 28.5 & 20.0 \\
\cline{2-4}
& DL \cite{dictionary} & 31.4 & 22.9 \\
\cline{2-4}
& HOG \cite{hog} & 34.2 & 25.7 \\
\cline{2-4}
& US-TL + HOG \cite{ijcbSkull} & 37.1 & 34.2 \\
\cline{2-4}
& SS-TL + HOG \cite{ijcbSkull} & 42.9 & 37.1\\
\hline
\hline
\multicolumn{2}{|l|}{\textbf{Proposed + HOG}} & \textbf{51.4} & \textbf{42.9}\\
\hline
\end{tabular}
\end{table}

\begin{itemize}
\item Table \ref{results} presents the rank-1 identification accuracy (\%)  of the proposed model on the IdentifyMe dataset for both the protocols, along with the comparative algorithms (averaged across five folds). It can be observed that with HOG features as input, the proposed Shared Transform Model achieves an accuracy of 51.4\% and 42.9\% on the two protocols, respectively.    

\item Shared Transform Model with HOG features showcases an improvement of almost 17\% as compared to the identification accuracy obtained with HOG features only. This demonstrates that the proposed model is able to encode discriminative features, useful for classification. Figure \ref{correct} presents sample skull and digital face image pairs correctly classified by the proposed model for Protocol-1. It is interesting to observe that the proposed Shared Transform Learning algorithm is able to model the large intra-class variations between data of different domains. 

\item As compared to the current state-of-the-art (SS-TL + HOG) \cite{ijcbSkull}, the proposed model presents an improvement of around 9\% and over 5\% for the first and the second protocol, respectively. An improvement of around 15\% is observed as compared to US-TL + HOG as well. 

\begin{figure}
\centering
\includegraphics[width = 2in] {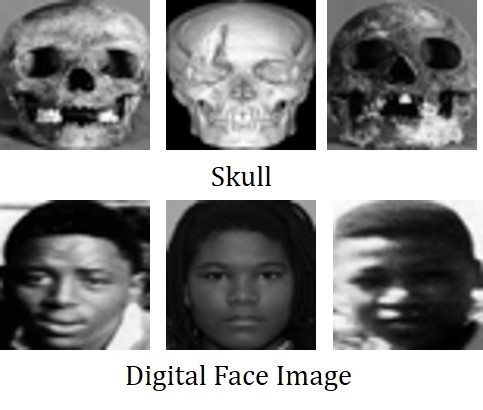}
\caption{Sample skull and digital face images correctly identified by the proposed Shared Transform Model. It can be observed that the proposed model is able to encode the heterogeneous variations present across both domains.}
\label{correct}
\vspace{-10pt}
\end{figure}

\item The focus of the proposed model is to learn a single transform matrix, whereas SS-TL learns two transforms and a classifier. The proposed model thus requires learning lesser number of parameters, which in turn results in improved performance for the given problem having limited training data. For an input of dimension $n\times1$, the proposed Shared Transform Model requires learning $n\times n$ parameters (shared transform matrix) for transforming the input to a dimension of $n\times1$. On the other hand, the SS-TL model requires learning $n\times n$ parameters for the skull transform, $n \times n$ for the digital face images transform, and  $n\times 1$ parameters for the perceptron based classifier. Therefore, the number of parameters learned in the proposed model is less than half as compared to SS-TL \cite{ijcbSkull}. We believe that given the small sample size problem of skull recognition, this reduction in the number of parameters improves the learning capability and generalizability of the model.  

\item Figure \ref{bar} presents the rank-5 identification accuracy (\%) of the proposed Shared Transform Model with HOG features on the IdentifyMe dataset. Accuracies pertaining to both the protocols are reported, where, the proposed Shared Transform Model outperforms other algorithms by reporting an improvement of at least 3\% as compared to the state-of-the-art results.

\begin{figure}
\centering
\includegraphics[width = 3in] {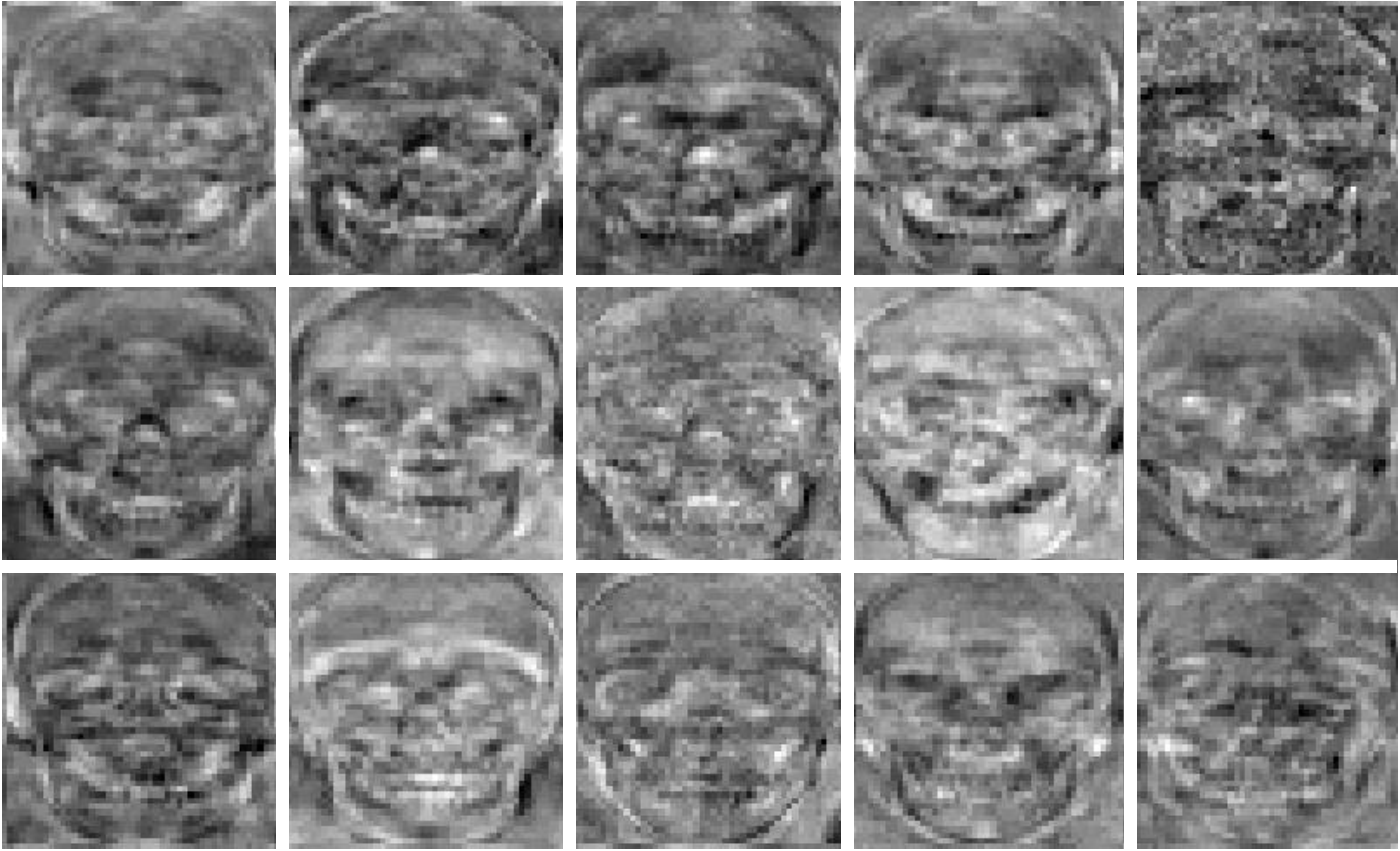}
\caption{Sample weights learned by the Shared Transform Model on raw pixels. It demonstrates that the model learns variations with respect to the structure of the input data.}
\label{weights}
\vspace{-10pt}
\end{figure}

\item Figure \ref{weights} presents sample weights learned by the proposed Shared Transform Model for the task of skull recognition, on raw pixel values. It is observed that the model encodes variations observed in different skull and digital face images. For instance, one can observe different overall structures, and variations specifically in the jaw and eye region of the weights. This demonstrates that the proposed model learns variables encompassing the variations observed in the input data. 

\item Upon analyzing the standard deviation of the algorithms across all folds, we observe that the proposed Shared Transform Model with HOG features presents a standard deviation of  7.1\% on Protocol-1. On the other hand, identification with HOG features results in a standard deviation of 11.9\%. For Protocol-2, the proposed model achieves a standard deviation of 6.3\%, whereas identification using HOG features results in a standard deviation of 23.4\%. The reduced standard deviation obtained across all folds for both the protocols showcases the robustness of the proposed approach for performing skull to digital face image matching. 
\end{itemize}

\section{Conclusion}
The task of skull to digital image matching is of vital importance for law enforcement in missing children cases, unsolved cold cases, or in instances of mass casualties. In all these real-world scenarios, the proposed algorithm can be valuable since it matches a skull image with gallery face images, if they exist. This can potentially reduce the manual processing burden on forensic artists and forensic anthropologists. The heterogeneous nature of the problem, along with the availability of limited training data, makes the problem challenging. In this research, a novel Shared Transform Model is proposed for performing skull to digital face image matching. The proposed model learns a ``shared transform'' for skulls and digital face images, while reducing the intra-class variations between the learned features. Experimental evaluation on the publicly available IdentifyMe dataset showcases the efficacy of the proposed model by achieving improved performance for the two protocols given with the dataset.  

\section{Acknowledgment}
This research is partially supported by Infosys Center for Artificial Intelligence at IIIT-Delhi. S. Nagpal is partially supported through TCS PhD Fellowship. 

{\small
\bibliographystyle{ieee}
\bibliography{submission_example}
}

\end{document}